\documentclass{article}

\usepackage{PRIMEarxiv}

\usepackage[utf8]{inputenc} 
\usepackage[T1]{fontenc}    
\usepackage{hyperref}       
\usepackage{url}            
\usepackage{booktabs}       
\usepackage{amsfonts}       
\usepackage{nicefrac}       
\usepackage{microtype}      
\usepackage{lipsum}
\usepackage{fancyhdr}       
\usepackage{graphicx}       
\usepackage{amsmath}
\usepackage{gensymb}

\graphicspath{{media/}}     

\title{Prediction of Maneuvering Status for Aerial Vehicles using Supervised Learning Methods}

\author{
  Abhishek Gupta \\
  Department of EXTC \\
  St. John College of Engineering and Management \\
  Palghar 401404, India\\
  \texttt{abhishekgupta@sjcem.edu.in} \\
   \And
  Sarvesh Thustu \\
  Department of EXTC \\
  St. John College of Engineering and Management \\
  Palghar 401404, India\\
  \texttt{sarvesht@sjcem.edu.in} \\
   \And
  Riti Thakor \\
  Department of EXTC \\
  St. John College of Engineering and Management \\
  Palghar 401404, India\\
  \texttt{ritit@sjcem.edu.in} \\
   \And
    Saniya Patil \\
    Department of EXTC \\
    St. John College of Engineering and Management \\
    Palghar 401404, India\\
    \texttt{saniya@sjcem.edu.in} \\
  \And
  Raunak Joshi \\
  Research Scholar \\
  University of Mumbai \\
  Mumbai 400032, India \\
  \texttt{raunakjoshi.m@gmail.com} \\
  \And
  Ronald Melvin Laban \\
  Department of EXTC \\
  St. John College of Engineering and Management \\
  Palghar 401404, India\\
  \texttt{ronaldl@sjcem.edu.in} \\
}

\begin{document}
\maketitle

\begin{abstract}
Aerial Vehicles follow a guided approach based on Latitude, Longitude and Altitude. This information can be used for calculating the status of maneuvering for the aerial vehicles along the line of trajectory. This is a binary classification problem and Machine Learning can be leveraged for solving such problem. In this paper we present a methodology for deriving maneuvering status and its prediction using Linear, Distance Metric, Discriminant Analysis and Boosting Ensemble supervised learning methods. We provide various metrics along the line in the results section that give condensed comparison of the appropriate algorithm for prediction of the maneuvering status.
\end{abstract}


\section{Introduction}
The maneuvers are very important metrics for aerial vehicle trajectories. The maneuvers measures the ability of the highly coordinated movements of aerial as well as ground vehicles. The aerial vehicles require this principle at a very important level and the detection of the maneuver can prove to be important for aerial simulations for trajectory calculation. Machine Learning can be leveraged for such type of problem, as finding the state of maneuvering is a binary classification \cite{10.5120/ijca2017913083} problem. Machine Learning in the area of supervised learning can help tackle this problem. Since supervised algorithms are taken into consideration, the constraint of using a parametric or non-parametric model is of not an utter importance. For classification we decided to consider a variety of algorithms. We targeted Linear \cite{10.2307/2344614}, Distance Measure \cite{Pandit2011ACS}, Discriminant Analysis \cite{Ramayah2010DiscriminantA} and Boosting Ensemble Method \cite{opitz1999popular}. The linear models have wide variety but decided to select Logistic Regression \cite{cramer2002origins,chung2020introduction} as it is the primordial algorithm used for binary classification problem. For distance measure we considered working with K-Nearest Neighbors \cite{10.1007/978-3-540-39964-3_62,9065747} which is a non-parametric supervised learning method. For discriminant analysis we considered Linear Discriminant Analysis \cite{salford52074,Li2014FisherLD,7952041,gupta2022discriminant} and for boosting ensemble method we considered CatBoost \cite{prokhorenkova2019catboost,dorogush2018catboost,Ibrahim2020,gupta2021succinct} algorithm. These algorithms are varied and will give varied results for maneuver prediction of trajectory based data.

\section{Methodology}
\subsection{Data Preparation}
The first thing is preprocessing the data in a conducive form for the machine learning algorithms. The required elements of the data typically focus on the Latitude, Longitude, Altitude and Timestamps of all the aerial vehicles travelling from one point to destination point. Such kind of data can be used for conversions required for one single data that works for algorithms. The maneuver can be calculated using many latitude, longitude and altitude. These are very influencing parameters for many features of dataset used in this project. Latitude gives measure of North-South location from the center of Earth and is measured in degrees. It symbol used for representing latitude is $\phi$. Equator is the very line drawn horizontally, that represents center of the Earth. Latitude is $0^{\degree}$ at Equator and $90^{\degree}$ at both the north and south pole. Longitude on the other-hand gives lines from north to south pole in a continuous manner. These lines are also called Meridians. These are denoted by $\lambda$ symbol. The prime meridian is calculated from Greenwich, England where it is considered as $0^{\degree}$ and runs all the way to $+/-180^{\degree}$, from East to West keeping track of Prime Meridian. The altitude is denoted by the $\alpha$ symbol and ranges from $0^{\degree}$ to $90^{\degree}$. The $n^{th}$ discrete difference is required for using Latitude, Longitude and Altitude for calculation of maneuver. Along with maneuver the parameters we require in data are maximum altitude, vertical acceleration, horizontal speed of plane and distance. These will make the major features for determining the maneuver factor. The formula for vertical acceleration is calculated using mean of $\alpha$, that is altitude. The data consists records of over 6000 files for calculation which have numerous coordinates which are calculated for parameters for dataset. The formula for horizontal speed of the plane is given by

\begin{equation}
    S_p = \sqrt{\phi^2+\lambda^2}
\end{equation}

This formula gives speed which then individually is multiplied with an arbitrarily initialized variable value that represents speed of the plane as per the standards. Later a mean of the values is taken to achieve horizontal speed of the plane. The distance can be calculated using

\begin{equation}
    D = \sqrt{(\phi_i-\phi_j)^2 + (\lambda_i-\lambda_j)^2}
\label{eq:b}
\end{equation}

The Equation \ref{eq:b} gives the distance and this is an important explanatory variable that is a feature influencing the maneuver. The maximum altitude is calculated by finding the maximum value of the $\alpha$. Finally we calculate the maneuver. First we arbitrarily denote a threshold for maneuver. The values falling above the maneuver are considered, others are labelled as zero. The formula can be denoted as

\begin{equation}
    M =
    \begin{cases}
       1, & \text{if}\ argmax{|\alpha|} > T \\
       0, & \text{otherwise}
    \end{cases}
    \label{eq:c}
\end{equation}

The Equation \ref{eq:c} states that Maneuver is calculated with a condition, the maximum absolute value of the $\alpha$ needs to be greater than declared Threshold $T$, in such case the maneuver is given as 1. If the condition is not followed it is 0. Using all the equations we calculated the values for over 6000 records of aerial vehicles and used this data for machine learning algorithms.

\subsection{Linear Model}
The Linear Model that we consider is Logistic Regression. It is linear parametric supervised learning method that uses a logit function for prediction. The logit function is smoothening curve based function which is also known as sigmoid. The formula for sigmoid can be given as

\begin{equation}
    \sigma(Z) = \frac{1}{1+e^{-Z}}
\end{equation}

where $Z$ is an equation that is parametric. This function is similar to Linear Regression function which is used for continuous value predictions. The Linear Regression function is enforced smoothness using sigmoid function for discrete value predictions. This function is given as

\begin{equation}
    Z = \beta_0 + (\beta_1*X_i) \nonumber
\end{equation}
where $\beta_0$ is regression constant and $\beta_1$ is regression coefficient representing one feature set.
\begin{equation}
    \sigma(Z) = \frac{1}{1+e^{\beta_0 + (\beta_1*X_i)}}
\end{equation}
This is a linear equation and later translates to many features as required and gives the dependent variable. it yields a threshold with certain probability ratio. This can be represented in formula as

\begin{equation}
    \hat{y} =
    \begin{cases}
       1, & \text{if}\ \hat{y} \geq T \\
       0, & \text{otherwise}
    \end{cases}
    \label{eq:d}
\end{equation}

The Equation \ref{eq:d} gives the representation prediction variable. $\hat{y}$ is considered as the prediction variable. If it is greater than or equal to the Threshold specified, it yields 1 else 0.

\subsection{Distance Measure Model}
The algorithm that we used for this type is K-Nearest Neighbor which is non-parametric supervised learning method. This algorithm takes an arbitrarily initialized K value which will be used to draw associations for all the data points from the data with respect to it. Usually the value of K is considered as odd value and more than 3. It then calculates the similarity value all the points under observation. This is done using some distance measure metric like Euclidean Distance. The formula for Euclidean is given as

\begin{equation}
    d(p, q) = \sqrt{\Sigma^n_{i=1}(p_i-q_i)^2}
\end{equation}

Similarly different types of distance metrics can be used. This help the algorithm determine the new sample based on K observation. Usually distribution of the data does not matter in Nearest Neighbors. Applied dimensionality reduction techniques for nearest neighbors sometimes help increase the accuracy of the algorithm.

\subsection{Discriminant Analysis}
The algorithm we considered for Discriminant Analysis is Linear Discriminant Analysis. These algorithms use Dimensionality Reduction and works for co-variance matrix. The estimates of the co-variance matrix are known as Scatter Matrix which is a semi definite matrix. These scatter matrices are of 2 types, Between Class Scatter and Within Class Scatter. The formula is given as

\begin{equation}
    LDA = \frac{(M_1-M_2)^2}{S^2_1+S^2_2}
\end{equation}

where numerator is between class scatter and denominator within class scatter. The equation needs to be expressed in the terms of W and needs to be maximized. After maximization of the numerator, the scatter matrix becomes $W^T.S_b.W$ and denominator becomes $W^T.S_w.W$. The equation is given as

\begin{equation}
    LDA = \frac{W^T.S_b.W}{W^T.S_w.W}
    \label{eq:e}
\end{equation}

Further the Equation \ref{eq:e} after differentiating with respect to W, eigenvalue and eigenvector \cite{denton2022eigenvectors} in a generalized manner is achieved.

\subsection{Boosting Ensemble}
The Ensemble Learning is a branch of Machine Learning that focuses on combining weak learners to yield a result equivalent or better than a strong learner. The boosting has major subdivisions known as Bagging \cite{10.1023/A:1018054314350} and Boosting. The bagging is known as Bootstrap Aggregation and Boosting uses the set of weak learners known as stumps and combines them to achieve the best possible result. The boosting methods include many algorithms out of which we considered Cat Boost. This is known as Categorical Boosting and is an advancement in the edition of boosting algorithms.

\section{Results}
\subsection{Precision}
This is a type of metric that defines the number of positive classes predicted from total number of positive classes. The foundation of the precision \cite{powers2020evaluation} starts from Confusion Matrix \cite{Ting2010ConfusionM}. The elements of confusion matrix are True Positives, True Negatives, False Positives and False Negatives. The formula for precision is given by

\begin{equation}
    Precision = \frac{TP}{TP+FP}
\end{equation}

The precision is again is has 2 types viz, Macro Averaging and Weighted Averaging. The macro averaging is where the precision does considers the unweighted averages. Whereas weighted averaging considers the every class while averaging.

\begin{table}[htbp]
\centering
\caption{Macro and Weighted Averaging Precision}\label{tab1}
\begin{tabular}{lll}
\toprule
Algorithm &  Macro & Weighted\\
\midrule
Logistic Regression &  85\% & 83\% \\
KNN &  81\% & 85\% \\
LDA & 82\% & 85\% \\
\textbf{CatBoost} & \textbf{85\%} & \textbf{88\%} \\
\bottomrule
\end{tabular}
\end{table}

The results from the Table \ref{tab1} give a detailed set of macro averaging and weighted averaging values for the all the algorithms used, but these values are not the fixed indication to derive a proper inference about the performance of the algorithm. For this we will have to include more metrics that will insights about learning process. Recall is the next metric we consider.

\subsection{Recall}
This is a metric that focuses on the positive predictions out of all the predictions. Recall \cite{powers2020evaluation} also works considering the elements of confusion matrix. The formula for recall is given by

\begin{equation}
    Recall = \frac{TP}{TP+FN}
\end{equation}

Just like precision, the recall also has the macro average and weighted average process.

\begin{table}[htbp]
\centering
\caption{Macro and Weighted Averaging Recall}\label{tab2}
\begin{tabular}{lll}
\toprule
Algorithm &  Macro & Weighted\\
\midrule
Logistic Regression &  66\% & 82\% \\
KNN &  79\% & 85\% \\
LDA & 78\% & 86\% \\
\textbf{CatBoost} & \textbf{81\%} & \textbf{88\%} \\
\bottomrule
\end{tabular}
\end{table}

The Table \ref{tab2} gives the metrics and it derives that Logistic Regression performs the worst. The indication is done using variation in macro and weighted averaging system. The least calculated variation for macro and weighted is in CatBoost.

\subsection{$F_1$-Score}
The $F_1$-Score \cite{powers2020evaluation} is a metric of accuracy that is calculated using Precision and Recall. The metric clearly gives assumption of the performance of the learning models. The formula for $F_1$-Score is given by

\begin{equation}
    F_1 = 2*\left(\frac{P*R}{P+R}\right) \nonumber
\end{equation}
this in terms of confusion matrix elements can be represented as
\begin{equation}
    F_1 = \frac{TP}{TP+\frac{1}{2}.(FP+FN)}
\end{equation}

Below is given a table that gives the $F_1$-Score for all the algorithms used, viz. Logistic Regression, K-Nearest Neighbors, Linear Discriminant Analysis and Cat Boost.

\begin{table}[htbp]
\centering
\caption{$F_1$-Score}\label{tab3}
\begin{tabular}{ll}
\toprule
Algorithm &  Score \\
\midrule
Logistic Regression & 82\% \\
KNN &  85\% \\
LDA & 86\% \\
\textbf{CatBoost} & \textbf{88\%} \\
\bottomrule
\end{tabular}
\end{table}

The $F_1$-Score gives the accuracy and as per the results CatBoost is able to give the best performance followed by LDA, followed by KNN and finally Logistic Regression.

\subsection{Receiver Operating Characteristics and Area Under Curve}

\begin{figure}[htbp]
\centering
\includegraphics[scale=0.7]{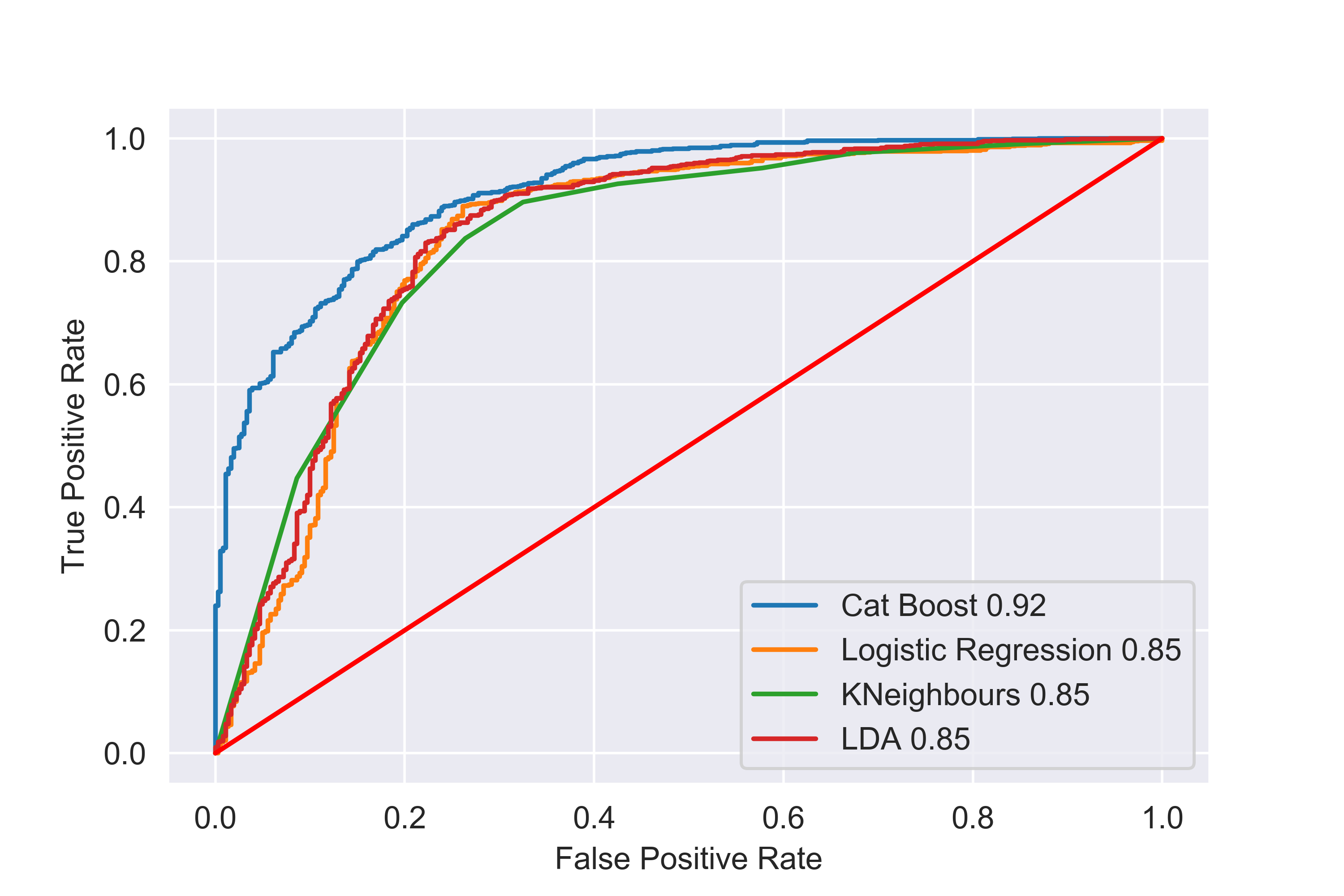}
\caption{RoC with AUC} \label{fig1}
\end{figure}

Receiver Operating Characteristics abbreviated as RoC \cite{Bradley1997TheUO} is performance measurement metric for classification models considering thresholds. It is formed using True Positive Rate and False Positive Rate \cite{vanRavenzwaaij2019TrueAF}. But the RoC is not sufficient, Area Under Curve abbreviated as AUC is essential for representing the measure of separability. The value of AUC is plot between 0 and 1. Closer the AUC to 1, better is the algorithm. The elements required for the graph need to be explained. True Positive Rate is also known as Sensitivity and also known as Recall. The formula is as same as that of Recall. For False Positive Rate, one cannot directly calculate the value of it. One needs to calculate Specificity first, which represented by the formula

\begin{equation}
    Specificity = \frac{TN}{TN+FP} \nonumber
\end{equation}

and now this can be used to find the False Positive Rate with formula as

\begin{equation}
    FPR = 1 - Specificity
\end{equation}

Now assuming this notion we have implemented the RoC and AUC for all the classifiers used in this paper. The Fig. \ref{fig1} gives the graphical representation of the classifiers used. The red diagonal is the threshold, any algorithm that falls below it is not considered a good algorithm. The inference can be drawn from the graph that CatBoost is the best classifier for the maneuver classification.

\section{Conclusion}
The data for aerial vehicles is allocated at a very high rate and can be definitely used to find the maneuver based on certain parameters. This is a binary classification problem and can be managed by leveraging Machine Learning. We wanted to provide the approach to tackle the problem in every single aspect of machine learning. Since it is binary classification, using Supervised learning methodology was straightforward approach, in which we considered working with Linear Model, Distance Metric Model, Discriminant Analysis and Boosting Ensemble Model. We used the best algorithm of each single approach that suits the best for our data. We used Logistic Regression for linear, KNN for distance metric, Linear Discriminant Analysis for discriminant analysis and CatBoost for boosting ensemble method. We provided the algorithms with metrics from different facets and proved the potential of every algorithm and successfully proved that maneuver detection for aerial vehicle data is possible. Obviously many more improvements in future can be done in future and we would be glad if someone else referred our work. 

\bibliographystyle{unsrt}  
\bibliography{references}

\end{document}